%% file: main.tex
\documentclass[%
 reprint,
 amsmath,amssymb,
 aip,
 prl,
]{revtex4-2}

\usepackage{dcolumn}

\usepackage[utf8]{inputenc}
\usepackage[T1]{fontenc}

\usepackage{graphicx}
\usepackage{bm}
\usepackage{comment}
\usepackage{algorithm,algpseudocode}
\usepackage{xcolor}
\usepackage{hyperref}
\usepackage{ulem}
\usepackage{enumitem}
\usepackage{comment}
\DeclareMathAlphabet{\mathcal}{OMS}{cmsy}{m}{n}
\newcommand{\E}[1]{ { \langle #1 \rangle } }

\renewcommand{\L}{\mathcal{L}}

\newcommand{\mytitle}{Intrinsic dimension estimation for discrete metrics}

\newcommand{\SI}{Supplementary Information}

\begin{document}
\preprint{AIP/123-QED}
\title{\mytitle}
\author{Iuri Macocco}
\affiliation{International School for Advanced Studies (SISSA), Via Bonomea 265,  34136 Trieste, Italy}
\author{Aldo Glielmo}
\affiliation{International School for Advanced Studies (SISSA), Via Bonomea 265,  34136 Trieste, Italy}
\affiliation{Banca d'Italia, Italy}
\altaffiliation{The views and opinions expressed in this paper are those of the authors and do not necessarily reflect the official policy or position of Banca d’Italia.}
\author{Jacopo Grilli}
\affiliation{The Abdus Salam International Centre for Theoretical Physics (ICTP), Strada Costiera 11, 34014 Trieste, Italy}
\author{Alessandro Laio}
\email{laio@sissa.it}
\affiliation{International School for Advanced Studies (SISSA), Via Bonomea 265,  34136 Trieste, Italy}
\affiliation{The Abdus Salam International Centre for Theoretical Physics (ICTP), Strada Costiera 11, 34014 Trieste, Italy}

\begin{abstract}
    
Real world-datasets characterized by discrete features are ubiquitous: from categorical surveys to clinical questionnaires, from unweighted networks to DNA sequences. Nevertheless, the most common unsupervised dimensional reduction methods are designed for continuous spaces, and their use for discrete spaces can lead to errors and biases. In this letter we introduce an algorithm to infer the intrinsic dimension (ID) of datasets embedded in discrete spaces. We demonstrate its accuracy on benchmark datasets, and we apply it to analyze a metagenomic dataset for species fingerprinting, finding a surprisingly small ID, of order 2. This suggests that evolutive pressure acts on a low-dimensional manifold despite the high-dimensionality of sequences' space.

    %
    %
    %
\end{abstract}

\maketitle

Data produced by experiments and observations are very often high-dimensional, with each data-point being defined by a sizeable number of features.
To the please of modelers, real-world datasets seldom occupy this high-dimensional space uniformly, as strong regularities and constraints emerge. 
Such a property is what allows for low-dimensional descriptions of these high-dimensional data, ultimately making science possible.

In particular, data-points are often effectively contained in a manifold which can be described by a relatively small amount of coordinates. 
The number of such coordinates is called Intrinsic Dimension (ID). 
%
More formally, the ID is defined as the minimum number of variables needed to describe the data without significant information loss.
Its knowledge is of paramount importance in unsupervised learning\cite{solorio2020review,jovic2015review,bengio2013representation} and has found applications across disciplines.
%
In solid state physics and statistical physics, the ID can be used as a proxy of an order parameter describing phase transitions\cite{Mendes-Santos2021,mendes2021intrinsic}; in molecular dynamics it can be used to quantify the complexity of a trajectory\cite{doi:10.1021/acs.chemrev.0c01195}; in deep learning theory the ID indicates how information is compressed throughout the various layers of a network \cite{ansuini2019intrinsic,doimo2020hierarchical,recanatesi2019dimensionality}.
During the last three decades much progress has been made in the development of sophisticated tools to estimate the ID\cite{Campadelli2015,Camastra2016} and
%
most estimators have been formulated (and are supposed to work) in spaces where distances can vary continuously. 
However, many datasets are characterised by discrete features and, consequently, discrete distances.
For instance, categorical datasets like satisfaction questionnaires, clinical trials, unweighted networks, spin systems, protein, and DNA sequences fall into this category.

Two main methods are usually employed in these cases.
The Box Counting (BC) estimator~\cite{falconer2004fractal,BCBlock,BCGrassberger}
--- which is defined by measuring the scaling between the number of boxes needed to cover a dataset and the boxes size --- provides good results for 2-3-dimensional datasets but is computationally demanding for higher-dimensional datasets.
The second popular method is the Fractal Dimension (FD) estimator\cite{falconer2004fractal,grassberger1983characterization,christensen2005complexity} and it is based on the assumption of a power law relationship $N\sim r^d$ for the number $N$ of neighbors within a sphere of radius $r$ from a given point, where $d$ is the fractal dimension of the data.
This estimator has been successfully applied, on discrete datasets, to model the phenomena of dielectric breakdown~\cite{niemeyer1984fractal} and Anderson Localization~\cite{kosior2017localization}.
%
For non-fractal objects, both methods are reliable only in the limit of small boxes and small radii, since the manifold containing the data can be curved and the data points can be distributed non-uniformly\cite{Facco2017}.
However, in discrete spaces such a limit is not well defined due to the minimum distance induced by any discrete lattice, and this can lead to systematic errors~\cite{Theiler:90,MOLLER1989176}.

In this letter, we introduce an ID estimator explicitly formulated for spaces with discrete features.
In discrete spaces, the ID can be thought of as the dimension of a (hyper)cubic lattice where the original data-points can be (locally) projected without a significant information loss. 
The key challenge in dealing with the discrete nature of the data lies in the proper definition of volumes on lattices.
To this end, we introduce a novel method that makes use of Ehrhart's theory of polytopes~\cite{ehrhart1977polynomes}, which allows to enumerate the lattice points of a given region.
By measuring a suitable statistics, depending on the number of data-points observed within a given (discrete) distance, one can infer the value of the dimension of the region, which we interpret as the ID of the dataset.
The statistics we use is defined in such a way that density of points is required to be constant only locally and not in the whole dataset. 
Importantly, our estimator allows to explicitly select the scale at which the ID is computed.
%

%
%
%
\textit{Methods} - 
We assume data points to be uniformly distributed on a generic domain, and that their density is $\rho$.
%
In such domain, we consider a region $A$ with volume $V\!(A)$.
%
Since we are assuming points to be independently generated, the probability of observing $n$ points in $A$ is given by the Poisson distribution~\cite{Moltchanov2012}
\begin{equation}
    P(n,A)=\frac{[\rho\, V\!(A)]^n}{n!}e^{-\rho\, V\!(A)}
    \label{eq:poisson}
\end{equation}
so that $\E n=\rho\, V\!(A)$.
Consider now a data-point $i$ and two regions $A$ and $B$, one containing the other, and both containing the data-point: $i\in A\subset B$. Then the number of points $n$ and $k-n$ falling, respectively, in $A$ and $B\setminus A$ are Poisson distributed with rates $\lambda_1=\rho \,V\!(A)$ and $\lambda_2=\rho\, V\!(B\setminus A)$. 
The conditional probability of having $n$ points in $A$ given that there are $k$ points in $B$ is
%
\begin{equation}
    P(n\,|\,k)=\frac{P(n)P(k-n)}{P(k)}=\binom{k}{n}p^n(1-p)^{k-n}
\end{equation}
with 
\begin{equation}
    p=\frac{\lambda_1}{\lambda_1+\lambda_2} = \frac{\rho\, V\!(A)}{\rho\, V\!(B)}= \frac{V\!(A)}{V\!(B)}.
    \label{eq:region_ratio}
\end{equation}
Thus $n|k\sim\mathrm{Binomial}(n;k,p)$. 
As far as the density $\rho$ is constant within $A$ and $B$, $p$ is simply equal to the ratio of the volumes of the considered regions and, remarkably, density independent.
This is a key property which, as we will show, allows using the estimator even when the density is approximately constant only locally, and varies, even substantially, across larger distance scales.
%
One can then write a conditional probability of the observations $n_i$ (one for each data point), given the parameters $k_i$ and $p_i$, which can possibly be point-dependent:
\begin{equation}
    \mathcal{L}(n_i|k_i,p_i)=\prod_{i=1}^N\mathrm{Binomial}(n_i|k_i,p_i).
    \label{eq:likelihood}
\end{equation}
Such formulation assumes all the observations to be statistically independent. Strictly speaking this is typically not true, since the regions $A$ and $B$ of different points can be overlapping.
We will address this issue in \SI~(SI), demonstrating that
neglecting correlations does not induce significant errors.

The next step consists in defining the volumes in Eq.~\eqref{eq:region_ratio} according to the nature of the embedding manifold.
We now assume our space to be a lattice where the $L^1$ metric is a natural choice. 
In this space the volume $V(A)$ is the number of lattice points contained in $A$.
%
According to Ehrhart theory of polytopes \cite{Ehrhart}, the number of lattice points within distance $t$ in dimension $d$ from a given point amounts to \cite{Beck2007}
\begin{equation}
    V_\diamond(t,d)= \binom{d+t}{d}\;_2F_1(-d,-t,-d-t,-1)
    \label{eq:enum_cross}
\end{equation}
where $_2F_1(a,b,c,z)$ is the ordinary hypergeometric function. At a given $t$, the above expression is a polynomial in $d$ of order $t$.
As a consequence, the ratio of volumes defining the value of $p$ in Eq.~\eqref{eq:region_ratio} becomes a ratio of two polynomials in $d$.
Given a dataset, the choice of $t_1$ and $t_2$ fixes the values of $n_i$ and $k_i$ in the likelihood~\eqref{eq:likelihood}. Its maximization with respect to $d$ allows to infer the data-manifold's ID,
%
which is simply given the root of equation (see SI for more details on the derivation)
\begin{equation}
   \frac{V_\diamond(t_1,d)}{V_\diamond(t_2,d)} - \frac{\E n}{\E k} = 0
    \label{eq:binom_mle}
\end{equation}
where the mean value over $n$ and $k$ is intended over all the points of the dataset. 
The root can be easily found with standard optimization libraries. This procedure defines an ID estimator that, for brevity, we will call I3D (Intrinsic Dimension estimator for Discrete Datasets).

%
Very importantly, the ID estimate is density independent as such factor cancels out (see Eq.~\eqref{eq:region_ratio}). 
%
%
The error on the estimator has a theoretical lower bound, given by the Cramer-Rao inequality, which has an explicit analytic expression.
As an alternative, the ID can be estimated by a Bayesian approach as the mean value of its posterior distribution, and the error estimated via the posterior variance (details in SI).
%

%
The estimation of the ID depends on the choice of the volumes of the smaller and larger regions, which are parametrised by the ``radii'' $t_1$ and $t_2$.
By varying $t_2$, the radius of the largest probe region, one can explore the behaviour of the ID at different scales.
The proper range of $t_2$ is dataset dependent and should be chosen by plotting the value of the ID as a function of it, as we will illustrate in the following. If the dataset has a well-defined ID, one will observe an (approximate) plateau in this plot.
This leaves the procedure with one free parameter: the ratio $r=t_1/t_2$ and its choice influences the statistical error. 
%
%
%
In continuous space the ratio between volumes in Eq.~\eqref{eq:region_ratio} is simply $p=r^d$ and the Cramer-Rao variance has a simple dependence on the parameter $r$.
By minimising it with respect to $r$, one obtains that the optimal value for the ratio is $r_{opt}\sim0.2032^\frac{1}{d}$ (see SI).

\begin{figure}[h]
    \centering
    \includegraphics[width=0.95\linewidth]{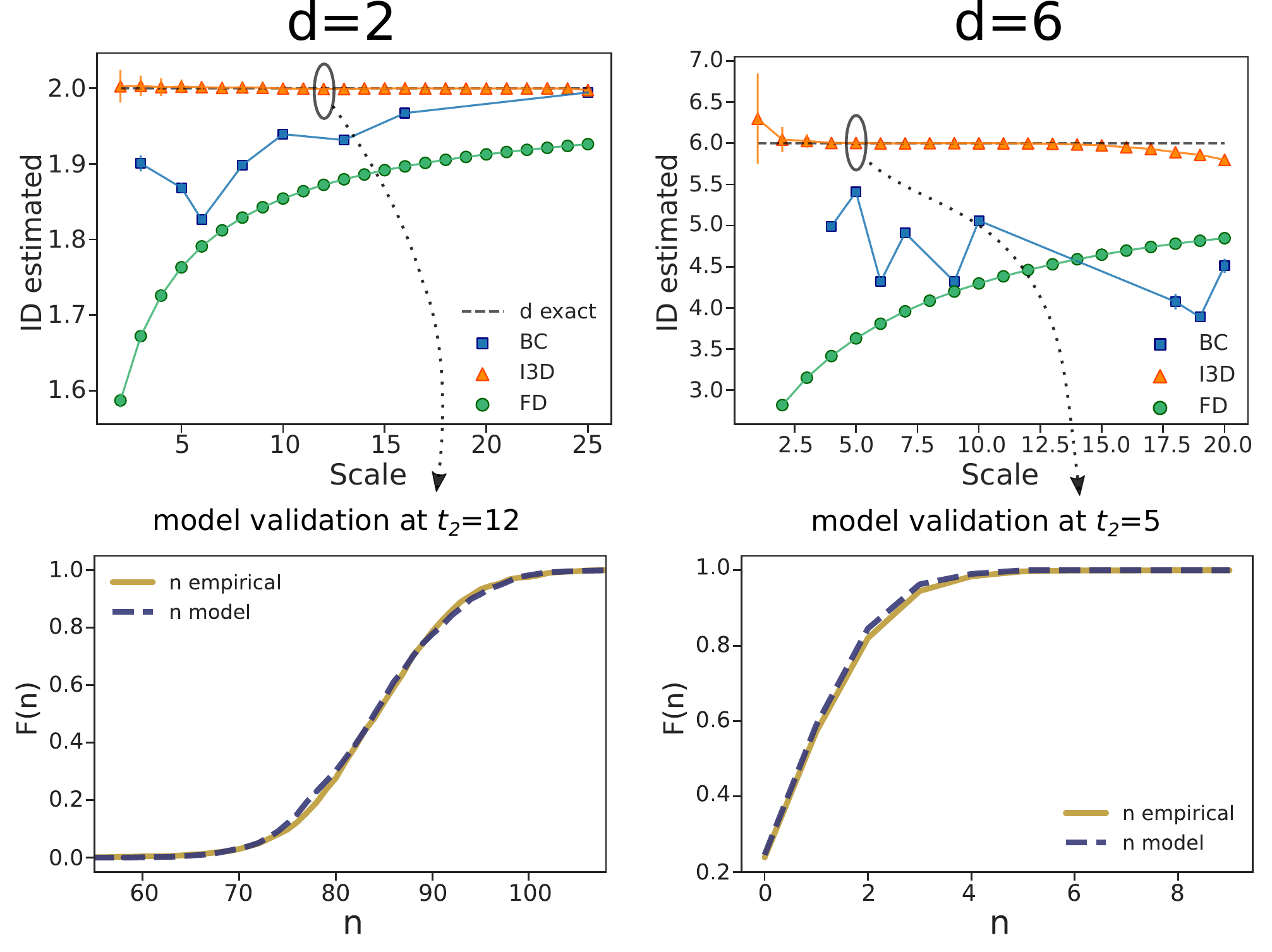}
    \caption{Performance of I3D, BC and FD estimators for points uniformly distributed on a square lattice of size 50 in 2d and and size 20 in 6d. Datasets were obtained by sampling, respectively, 20 realizations of 2500 and 100000 points. Error bars are given by the standard deviation over the different realizations.
    Lower panels: I3D model validation performed by comparing empirical and theoretical cdfs of the random variable $n$. 
    }
    \label{fig:uniform}
\end{figure}

In order to check the goodness of the estimator, we test whether the number of points $n$ contained within the internal shells are actually distributed as a mixture of binomials, as our model assumes:
\begin{equation}
    P(n)=\sum_kP(k)\mathrm{B}(n;k,\frac{V_\diamond(t_1,d)}{V_\diamond(t_2,d)})
    \label{eq:theoret}
\end{equation}
where 
$P(k)$ is the the empirical probability distribution of the $k$ found by fixing $t_2$. 
In the following we will compare the empirical cumulative distribution of $n$ to the cumulative distribution of $P(n)$.
\textit{Results: Uniform distribution} - We tested the I3D estimator on artificial datasets, and compared it against the two aforementioned methods: the Box Counting (BC) and the Fractal Dimension (FD). The BC estimate of the ID is obtained by a linear fit between the logarithm of the number of occupied covering boxes and the logarithm of the boxes side. Seemingly, for the FD, the linear fit is computed among the logarithm of the average number of neighbours within a given radius and the logarithm of the radius. In both cases, the scale reported in the figures is given by the largest box or radius included in the fit.
We started by analysing uniformly-distributed points in 2d and 6d square lattices. We adopted periodic boundary conditions in order to reduce boundary effects as much as possible. For the I3D estimator, in this and all following cases, we set $t_1/t_2=r=0.5$. Results are shown in Fig.~\ref{fig:uniform}.
While the BC and FD proved to be reliable in finding the fractal dimension of repeating, self-similar lattices~\cite{falconer2004fractal,niemeyer1984fractal}, they do not manage to assess the proper dimension of randomly distributed points, especially at small scales.
The I3D estimator, instead, returns accurate values for the ID at all scales and, importantly, provides the correct estimate also on self-similar lattices (see SI).
%
Remarkably the I3D estimator allows to select the scale explicitly by varying the radius $t_2$. 
In Fig. \ref{fig:uniform}, lower panels, we also report a first example of model validation for I3D.
The two cdfs (empirical and theoretical one, according to Eq.~\eqref{eq:theoret}) perfectly match, meaning that the ID estimation is reliable.

\textit{Gaussian distribution} - Secondly, we tested the estimators on Gaussian distributed points in 5 dimensions, analysing a case in which the data are uncorrelated and a case in which a correlation is induced by a non-diagonal covariance matrix. In both cases, we set diagonal elements of the covariance matrix to $\sigma=5$ (implying an effective standard deviation of the distribution of $\sigma_{eff}=\sqrt{d}\sigma$), while off-diagonal terms --for correlated data-- were uniformly extracted in the interval (0,2). The values were chosen in order to keep the dimension of the dataset under control, as correlations of the same order of the diagonal would reduce the dimensionality of the dataset.
The points were projected on a lattice by taking the nearest integer in each coordinate.
\begin{figure}[h]
    \centering
    \includegraphics[width=1\linewidth]{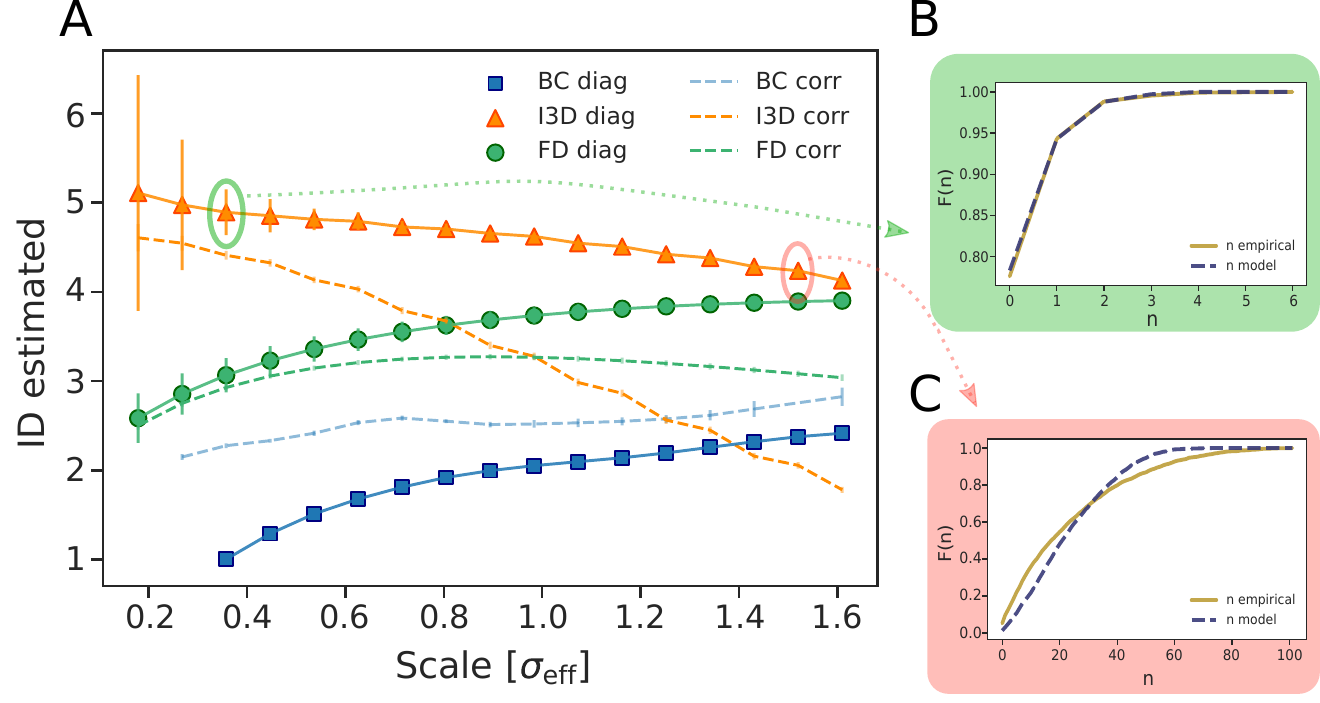}
    \caption{ID estimations of I3D, BC and FD on 20 realizations of 2500 points drawn from a Gaussian distributions in 5d and projected on a lattice (\textbf{A}). Solid lines with markers are related to diagonal covariance matrix, dashed lines to the non-diagonal case.
    Panels \textbf{B} and \textbf{C} show, respectively, I3D model validation at a small and large scales. 
    }
    \label{fig:gauss}
\end{figure}
As one can observe in panel \textbf{A} of Fig.\ref{fig:gauss}, I3D is accurate as far as it explores a neighborhood where the density does not vary too much (namely, as far as $t_2/\sigma_{eff}\lesssim1$). Correspondingly, empirical and model cdfs in panel \textbf{B} are superimposed.
Beyond such distance, neighborhoods are characterised by non-constant density; consequently, estimates gets less precise and, accordingly, the two cdfs show inconsistencies (panel \textbf{C}: $t_2/\sigma_{eff}\sim1.5$). 
On the other hand, the BC and FD estimations are far from desired values at any scales, for both correlated and uncorrelated cases.

\begin{figure}[htb]
    \centering
    \includegraphics[width=0.95\linewidth]{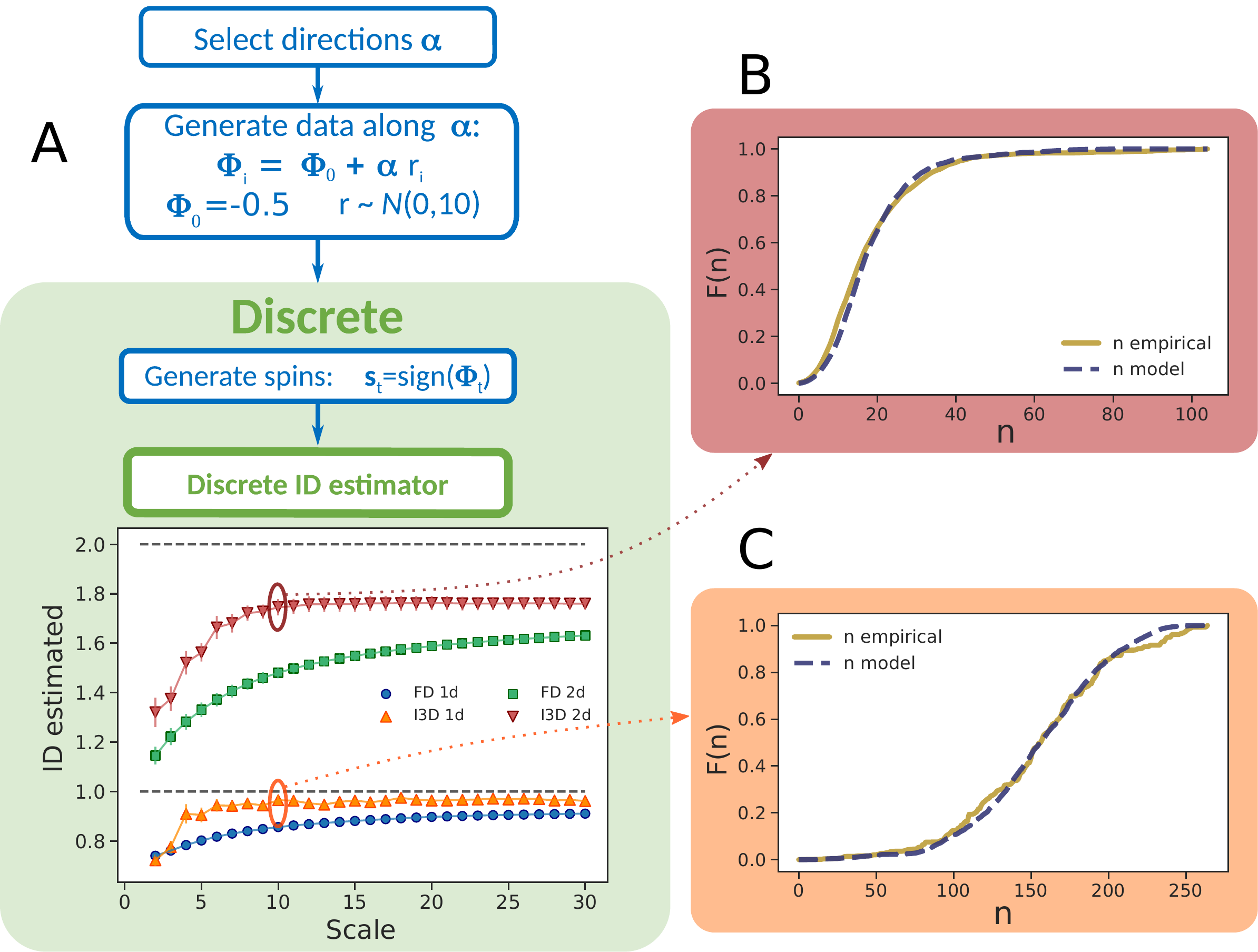}
    \caption{(\textbf{A}) The pipeline used to create an ensemble of binary spins with a low ID, together with the results of FD and I3D estimators on 1d and 2d datasets. 
    I3D estimations were validated by comparing theoretical and empirical cdfs (panels \textbf{B} and \textbf{C}). 
    }
    \label{fig:ising}
\end{figure}

\textit{Spin dataset} - 
As a third test, we created synthetic Ising-like spin systems with a tunable ID, which is given by the number of independent parameters used to generate the dataset. The 1d ensemble is obtained by generating a set of points belonging to a line embedded in $\mathbb{R}^D$ with the process $\bm{\varphi}_i=\bm{\varphi}_0+\bm\alpha\epsilon(i)$. Here, $\bm\alpha$ is a fixed random vector of unitary norm with uniformly distributed components and $\bm{\varphi}_0=-0.5$ is the y-intercept that, for simplicity, is equal for all the components; $\epsilon_i$ are gaussian-distributed: $\epsilon\sim\mathcal{N}(0,10)$ and independently drawn for each sample $i$.
We then proceed to the discretization by extracting the $\bm{z}_i=\text{sign}(\bm{\varphi}_i)$, an ensemble of $N$ states of $D$ discrete spins. The pipeline is summarized in Fig.\ref{fig:ising}. The role of $\bm{\varphi}_0$ is to introduce an offset in order to enhance the number of the reachable discrete states. In fact, for $\bm{\varphi}_0=0$, we would obtain only two different states, given by $\bm{z}=\text{sign}(\bm\alpha\epsilon)=\pm\text{sign}(\bm\alpha)$, since the spins would change sign synchronously. An offset $\ne0$ allows the angles $\bm{\varphi}_i$ and the spins $\bm{z}_i$ to shift sign in an asynchronous way.
The extension to higher dimensions is straightforward and consists in generating the initial points as $\bm{\varphi}_i=\bm{\varphi}_0+\sum_{j=1}^{id}\bm\alpha_j\epsilon_j(i)$, with $\bm\alpha_j\cdot\bm\alpha_k\sim\delta_{jk}$. 
Due to the nature of data domain (a $D$-dimensional hypercube with side 1), the BC cannot be applied, as boxes with side larger than 1 would include the whole data set.
FD and I3D estimates for the 1d system are very close. This is not surprising as both continuous and discrete volumes (and, consequently, the neighbors) scale linearly with the radius. 
%
%
In the 2d case, I3D clearly outperforms the other methods, although even the best estimate remains slightly lower than the true value.
This effect, due to non-uniform density, is relatively small and indeed the empirical and theoretical cdfs are rather consistent (panels \textbf{B} and \textbf{C}).
Such an effect becomes more important as the dimension rises (see SI for examples in $d=3$ and $d=4$).

\textit{16S Genomics strands} - Lastly, we present the application of our methodology to a real-world dataset in the field of genomics.
The dataset consists of DNA sequences of $\sim\!100\!-\!300$ nucleotides.
We selected a dataset downloaded from the Qiita server (https://qiita.ucsd.edu/study/description/13596)\cite{framework|qiime2:2020.11.1|0}. In such study, they sequenced the v4 region of the 16S ribosomal RNA of the microbiome associated with sponges and algal water blooms.
This small-subunit of rRNA genes is widely used to study the composition of microbial communities \cite{gray1984evolutionary,woese1990towards,weisburg199116s,jovel2016characterization}. 
Hamming distance and the binary mapping A:11, T:00, C:10 and G:01 were used to compute sequences' distance. The canonical letter representation leads to almost identical results (see SI).
To avoid dealing with isolated sequences, we kept only sequences having at least 10 neighbors within a distance of 10. 
Sequences come with their associated multiplicity, related to the number of times the same read has been found in the samples. We ignore such degeneracy and compute an ID which 
describes just the distribution of the points regardless of their abundance.

\begin{figure}[htb]
\centering
\includegraphics[width=0.95\linewidth]{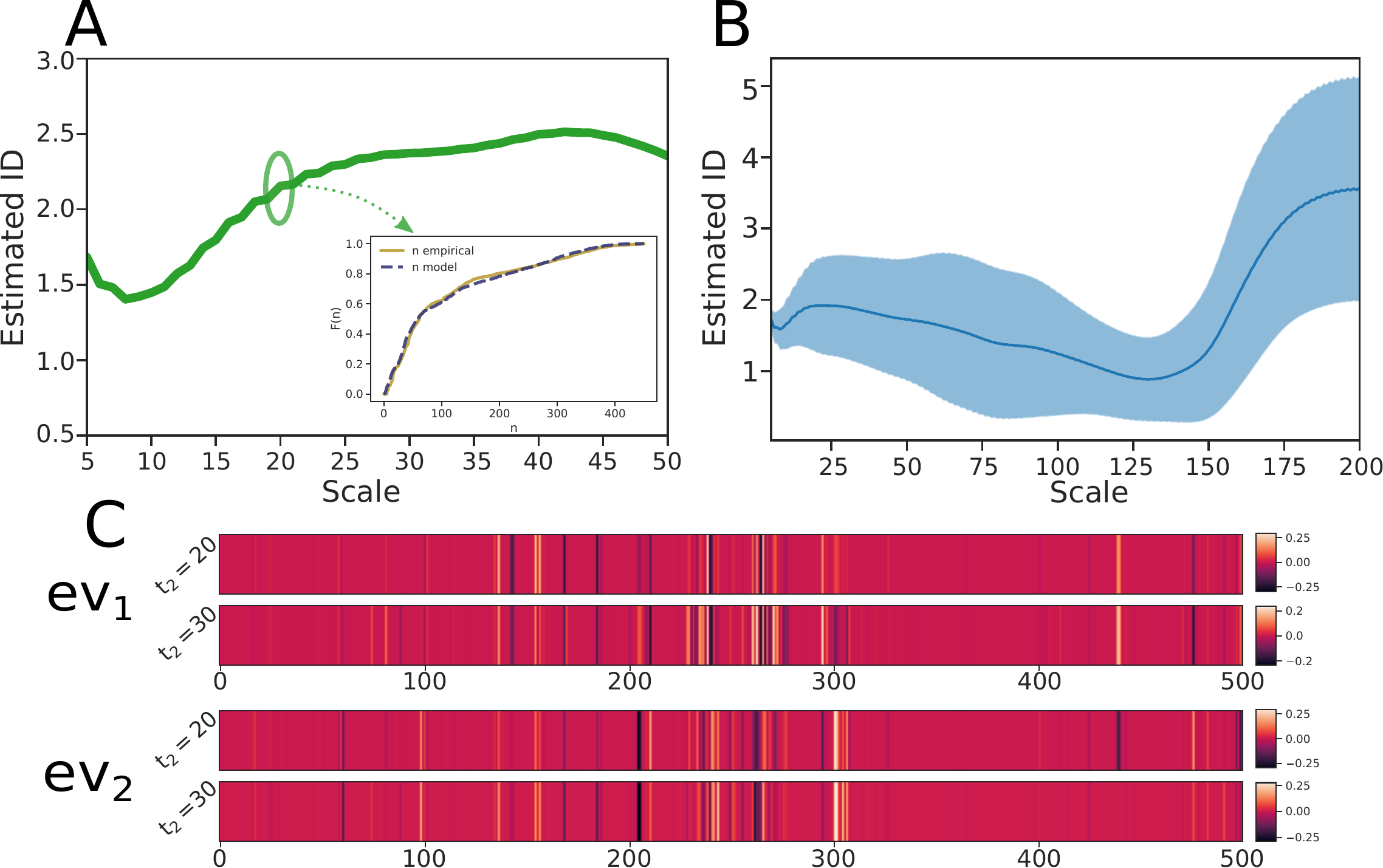}
\caption{Estimated ID at small to medium distances for one of the clusters of the genomics dataset (panel \textbf{A}). The inset reports the fair superposition of empirical and modeled cdfs of $n$. Panel \textbf{B} shows average and standard deviation of the IDs estimated separately for each cluster.
Panel \textbf{C} shows first and second PCA eigenvectors of the data-points within given distances $t_2$ (20 or 30) from the center of cluster used for panel \textbf{A}. 
}
\label{fig:fig4}
\end{figure}

To begin with, we estimated the ID on a subset of sequences that are similar to each other. In order to find such sets, we perform a k-means clustering and calculate the ID separately for each of them.
Panel \textbf{A} in Fig. \ref{fig:fig4} shows the ID at small to medium scale for one of such clusters.
The empirical and reconstructed cdfs, performed at $t_2=20$ (see inset), are fairly compatible.
Panel \textbf{B} shows the average and the standard deviation of the ID of all clusters (weighted according to the respective populations). One can appreciate that the ID is always between 1 and 3 in a wide range of distances, showing a plateau around 2 for $15<t_2<40$. 

Such a low value for the ID is an interesting and unexpected feature, as it suggests that, despite the high-dimensionality of sequences' space, evolution effectively operates in a low-dimensional space. 
Qualitatively, an ID $\sim$ 2 on a scale of $\sim$ 20 means that if one considers all the sequences differing by approximately 20 mutations from a given sequence, these mutations cannot be regarded as independent one from each other, but are correlated in such a way that approximately 18 degrees of freedom are effectively forbidden.
The ``direction'' of these correlated mutations can be, at least approximately, measured by performing PCA in the space of sequences with the binary mapping.
The first two dominant eigenvectors, shown in panel \textbf{C}, were estimated using all the sequences within a distance of 20 (top) and 30 (bottom) from the center of the cluster of Panel \textbf{A}. Remarkably, the eigenvectors do not change significantly on this distance range, indicating that, consistently with the low value of the ID, the data manifold on this scale can be approximately described by a two-dimensional plane.
In order to provide an interpretation of the vectors defining this plane, we repeated this same analysis on the previously mentioned spin model. In this case, if the generative model is defined by two vectors  $\bm \alpha_1$ and $\bm \alpha_2$, the first two dominant eigenvectors of a PCA performed on $\sim$1000 points are contained in the span of the two generating vectors, with a residual of 0.04 (see SI for details).   
The components of a vector $\bm \alpha$ can then be qualitatively interpreted as proportional to the mutation probabilities of the associated nucleotide for a collective mutation process.
In the genomics dataset this reasoning can applied only locally: the direction of correlated mutation is significantly different in different clusters, indicating that the data manifold is highly curved.
\textit{Conclusions} - We presented an ID estimator formulated to analyze discrete datasets. Our method relies on few mathematical hypotheses and is asymptotically correct if the density is constant within the probe radius $t_2$. 
In order to prove the estimator's effectiveness, we tested the algorithm against three different artificial datasets and compared it to the well known Box Counting and Fractal Dimension estimators. While the last two performed poorly, the new one achieved good results in all cases, providing reliable ID estimations corroborated by the comparison of empirical and model cumulative distribution functions for one of the observables.
We finally applied the estimator on a genomics dataset, finding an unexpectedly low ID which hints at strong constraints in the sequences' space, and then exploited such information to give a qualitative interpretation of such ID.
The newly developed method paves the way to push the investigation even further, towards the extension to discrete metrics of distance-based algorithms and routines that are, nowadays, consolidated in the continuum, such as density estimation methods or clustering algorithms.\\

The code implementing the algorithm is available in open source within the DADApy \cite{dadapy_final} software.

\section{Acknowledgements}
The authors thank Antonietta Mira, Alex Rodriguez and Marcello Dalmonte for the fruitful discussions. AG and AL acknowledge support from the European Union’s Horizon 2020 research and innovation program (Grant No. 824143, MaX ‘Materials design at the eXascale’ Centre of Excellence).\\

IM, AG, AL designed and performed the research. All authors wrote the paper. JG designed the application on genomics sequences.

\bibliography{IDD_paper}
\newpage
\include{SI}
\end{document}

%% file: SI.tex
\onecolumngrid
\begin{center}
\textbf{\large{\mytitle\\--- Supplementary Information ---}}
\end{center}

\begin{figure}[h!]
    \centering
    \includegraphics[width=0.7\linewidth]{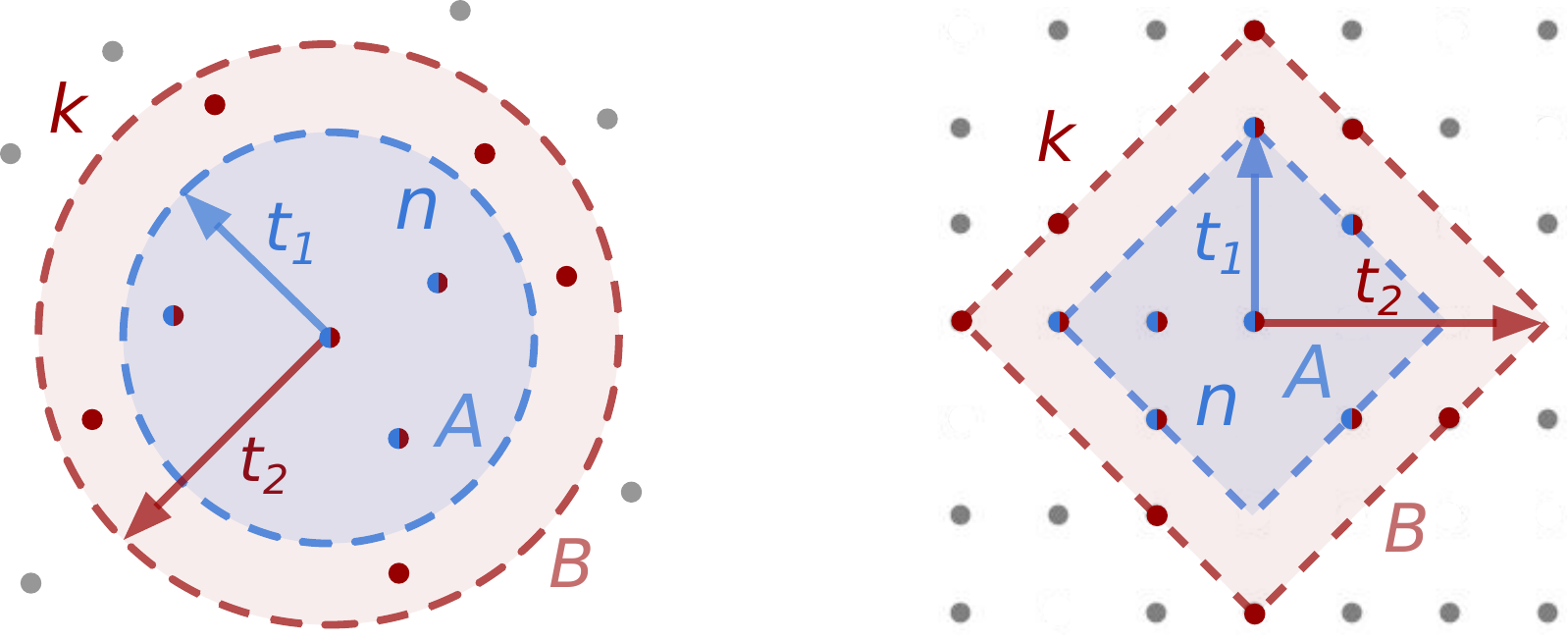}
    \caption{Sketchy representation of hyperspheres for the typical $L^2$ metric in continuous spaces (left) and for lattices (right).
    In order to find the ID we exploit the binomial relationship between the $n$ (blue) points within region $A$ -of radius $t_1$- and $k$ (red) points within region $B$ -of radius $t_2$-. Points within the inner region $A$ count for both $n$ and $k$.}
    \label{fig:estimator}
\end{figure}
\section{Proof of Eq.(2)}
We recall that
\begin{equation}
\begin{split}
        P(n,A)&=\frac{[\rho\, V\!(A)]^n}{n!}e^{-\rho\, V\!(A)}\\
        P(k,B)&=\frac{[\rho\, V\!(B)]^k}{k!}e^{-\rho\, V\!(B)}\\
        P(k-n,V(B\setminus A))&=\frac{[\rho\, V\!(B\setminus A)]^{k-n}}{(k-n)!}e^{-\rho\, V\!(B\setminus A)}
\end{split}
\end{equation}
where we name $\lambda_1=\rho \,V\!(A)$ and $\lambda_2=\rho\, V\!(B\setminus A)$, so that $\lambda_1+\lambda_2=\rho \,(V\!(A)+V\!(B\setminus A))=\rho \,V\!(B)$ since $A\subset B$.
Consequently, it holds that
\begin{equation}
\begin{split}
    P(n\,|\,k)&=\frac{P(n,k)}{P(k)}=\frac{P(n)P(k-n)}{P(k)}=\\
    &=\binom{k}{n}\left(\frac{\lambda_1}{\lambda_1+\lambda_2}\right)^n\left(\frac{\lambda_2}{\lambda_1+\lambda_2}\right)^{k-n}=\\
    &=\binom{k}{n}p^n(1-p)^{k-n}
\end{split}
\end{equation}
where $p=\frac{V(A)}{V(B)}$.

\section{Ehrhart polynomial theory and Cross-polytope enumerating function}
According to Ehrhart theory, the volume of a lattice hypershpere of radius $t$ in $d$ dimension is given by the enumerating function\cite{Beck2007}
\begin{equation}
    V_\diamond(t,d)= \sum_{k=0}^d\binom{d}{k}\binom{t-k+d}{d}.
 \label{eq:supp1}
\end{equation}
where $d$ is assumed to be an integer value. In order to make this expression suitable for likelihood maximization, it can be conveniently rewritten using the analytical continuation
\begin{equation}
    V_\diamond(t,d) = \binom{d+t}{d}\;_2F_1(-d,-t,-d-t,-1)%
\end{equation}
where the binomial coefficient are computed using the Gamma function: $n!=\Gamma(n+1)$ for non-integer $n$.
$_2F_1$ is the hypergeometric function. Here we report the first polynomials for $t\le4$:
\begin{itemize}[leftmargin=2.0cm]
    \item[$t=0$:] 1
    \item[$t=1$:] $1+2d$
    \item[$t=2$:] $1+2d+2d^2$
    \item[$t=3$:] $1+\frac{8}{3}d+2d^2+\frac{4}{3}d^3$
    \item[$t=4$:] $1+\frac{8}{3}d+\frac{10}{3}d^2+\frac{4}{3}d^3+\frac{2}{3}d^4$
\end{itemize}
By substituting integer values of $d$, one recovers the volumes found with eq. \ref{eq:supp1}. Using  this expansion we can treat $d$ as a continuous parameter in our inference procedures.

\section{Maximum Likelihood Estimation of the ID}
\label{sec:MLE}
Once the two radii $t_1$ and $t_2$ are fixed and the corresponding values of $n_i$ and $k_i$ are computed, the likelihood for $N$ data points is
\begin{equation}
    \mathcal{L}\{d|(n_i,k_i)\}=\prod_i^N\mathrm{Binomial}(n_i;k_i,p(d))=\prod_i^N\binom{k_i}{n_i}(p(d))^{n_i}(1-p(d))^{k_i-n_i}
    \label{eq:L}
\end{equation}
where $p(d)=V_\diamond(t_1,d)/V_\diamond(t_2,d)$ and depends explicitly only on $d$. In order to make the expressions easier to read, we will write just $p$ from now on. The optimal value for $d$ can be found by means of a maximum likelihood estimation (MLE), which consists in setting equal to 0 the (log)derivative of the likelihood:
\begin{equation}
\begin{split}
    0 =&\frac{\partial}{\partial d}\ln(\L)=\sum_i^N\frac{\partial}{\partial d}\ln(\mathrm{B}(n_i;k_i,p))=\\
    =&\sum_i^N\frac{\partial}{\partial d}\left(n_i\ln(p)+(k_i-n_i)\ln(1-p)\right)=\\
    =&\sum_i^N\left(n_i\frac{p'}{p}-(k_i-n_i)\frac{p'}{1-p}\right)=\\
    =&\E n\frac{p'}{p}-(\E k- \E n)\frac{p'}{1-p}
\end{split}
\label{eq:MLE_1}
\end{equation}
where the mean value are intended over all the points of the dataset and $p' = \text{d} p/\text{d} d$ is the Jacobian of the transformation from $p$ to $d$, which reads
\begin{equation}
    \frac{\text{d}}{\text{d}d}\frac{V_\diamond(t_1;d)}{V_\diamond(t_2;d)}=\frac{V_\diamond'(t_1)V_\diamond(t_2)-V_\diamond(t_1)V_\diamond'(t_2)}{V_\diamond(t_2)^2}.
\label{eq:discrete_jacobian}
\end{equation}
The last line of Eq.~\eqref{eq:MLE_1} leads directly to equation~\eqref{eq:binom_mle}. 
\subsection{Cramer-Rao estimate of the variance of the ID}
\label{sec:CR}
The Cramer-Rao inequality states that the variance of an unbiased estimator of an unknown parameter is bounded from below by the inverse of the Fisher information, namely
\begin{equation}
    \text{Var}(\hat{\theta}) \ge \frac{1}{\mathcal{I}(\theta)}
\end{equation}
where
\begin{equation}
    \mathcal{I}(\theta) = N\mathbb{E}\left[\left( \frac{\partial}{\partial\theta}\ln\L(x;\theta)\right)^2\right]=-N\mathbb{E}\left[\frac{\partial^2}{\partial\theta^2}\ln\L(x;\theta)\right],
\end{equation}
$\L$ is the likelihood for a single sample $x$ and $\mathbb{E}$ is the expected value over datapoints $x$.
Given the likelihood in eq.~\eqref{eq:L}, by deriving (with respect to $d$) a second time the third line of eq.\eqref{eq:MLE_1}, one has
\begin{equation}
    \frac{\partial^2}{\partial d^2}\ln\mathcal{L}(d;n_i)=\sum_i^N\left( p''(\frac{n_i}{p}-\frac{k_i-n_i}{1-p})+p'(-p'\frac{n_i}{p^2}-p'\frac{k_i-n_i}{(1-p)^2})\right).
    \label{eq:CR_explicit}
\end{equation}
By inserting the MLE solution $\E n=p\E k$ and performing the, sum one obtains
\begin{equation}
    \frac{\partial^2}{\partial d^2}\ln\mathcal{L}(d;n_i) = -N\E k \frac{p'^2}{p(1-p)},
    \label{eq:CR_discrete}
\end{equation}
leading to the final result 
\begin{equation}
    \text{Err}(d;t_1,t_2,N) \ge \sqrt{\frac{p(1-p)}{\E k N p'^2}}\Bigg|_{d=d_{MLE}}.
    \label{eq:CR_var}
\end{equation}
Such an expression is finally evaluated at the $d$ found through the MLE procedure.
\section{Bayesian estimate of the ID}
\label{sec:bayes}
The  ID can also be estimated through a Bayesian approach. The likelihood of the process is represented by a binomial distribution where the probability $p$ (hereafter named $x$ to avoid confusion) is given by the ratio of the shell volumes around any given point. The binomial pdf has a known conjugate prior: the beta distribution, whose expression is
\begin{equation}
\text{Beta}(x;\alpha,\beta)=\frac{x^{\alpha-1}(1-x)^{\beta-1}}{\mathcal{B}(\alpha,\beta)},
\label{eq:beta_dist}
\end{equation}
where $\mathcal{B}$ is the beta function.
We make an agnostic assumption (since with do not have any information on the value of the ID) and set $\alpha=\beta=1$, so that $x$ will be uniformly distributed.
The posterior distribution of the ratio of the volumes $x$ will still be a Beta distribution, with parameters updated as follows
\begin{eqnarray}
     \alpha &=& \alpha_0 + \sum_{i=1}^Nn_i\\
     \beta &=& \beta_0 + \sum_{i=1}^N(k_i-n_i)
\label{eq:beta_par}
\end{eqnarray}
where $n_i$ and $k_i$ are the points falling within the inner and outer volumes around point $i$; the sum runs on all the points in the dataset. 

In order to compute the expected value and the variance on $d$, one has to extract its posterior distribution $P(d)$ from the one of $x$.
The posterior of $d$ is obtained from the posterior of $x$ by a simple change of variables: 
\begin{equation}
    P(d) = P(p(d))\left|\frac{\text{d}p}{\text{d}d}\right|=\text{Beta}(p;\alpha,\beta)\left|\frac{\text{d}p}{\text{d}d}\right|
    \label{eq:posterior}
\end{equation}
where the Jacobian is given by Eq.~\eqref{eq:discrete_jacobian}.
By varying $d$, one can estimate the posterior distribution of $d$. Its first and second momenta will be the (Bayesian) estimates of the ID and of its confidence.

As far as we could observe, the ID found through MLE is always very close to the mean value of the posterior. The same occurs for the error estimate, which is typically very close to the Cramer-Rao  bound. Small differences ($<1\%$) have been observed in cases of few datapoints ($\sim50$). The reason is that the posterior distribution, for low values of $\alpha$ and $\beta$, can be slightly asymmetric, bringing to a discrepancy between the position of its maximum and its mean value. In most practical cases such an effect is negligible.
\section{Statistical correlation in the numbers of data points in the probe regions.}
\label{sec:stat_indep}
The likelihood has the form in Eq. \ref{eq:L} under the assumption that the observations $n_i$ are statistically uncorrelated. 
However, if a point $j$ is close to another point $i$, it is likely that their neighborhoods will be overlapping. This implies that the values for $n_i$ and $n_j$ will have a certain degree of correlation. In particular, in order to have a fully statistical independence, one should consider only points with a non-overlapping neighbours. Clearly this would reduce the number of available observations to compute the ID, making the estimate less reliable. 
Here we assess the entity of such correlations, or seemingly, how much the Bayesian and Cramer-Rao calculations underestimate the error.

\begin{figure}
    \centering
    \includegraphics[width=0.65\linewidth]{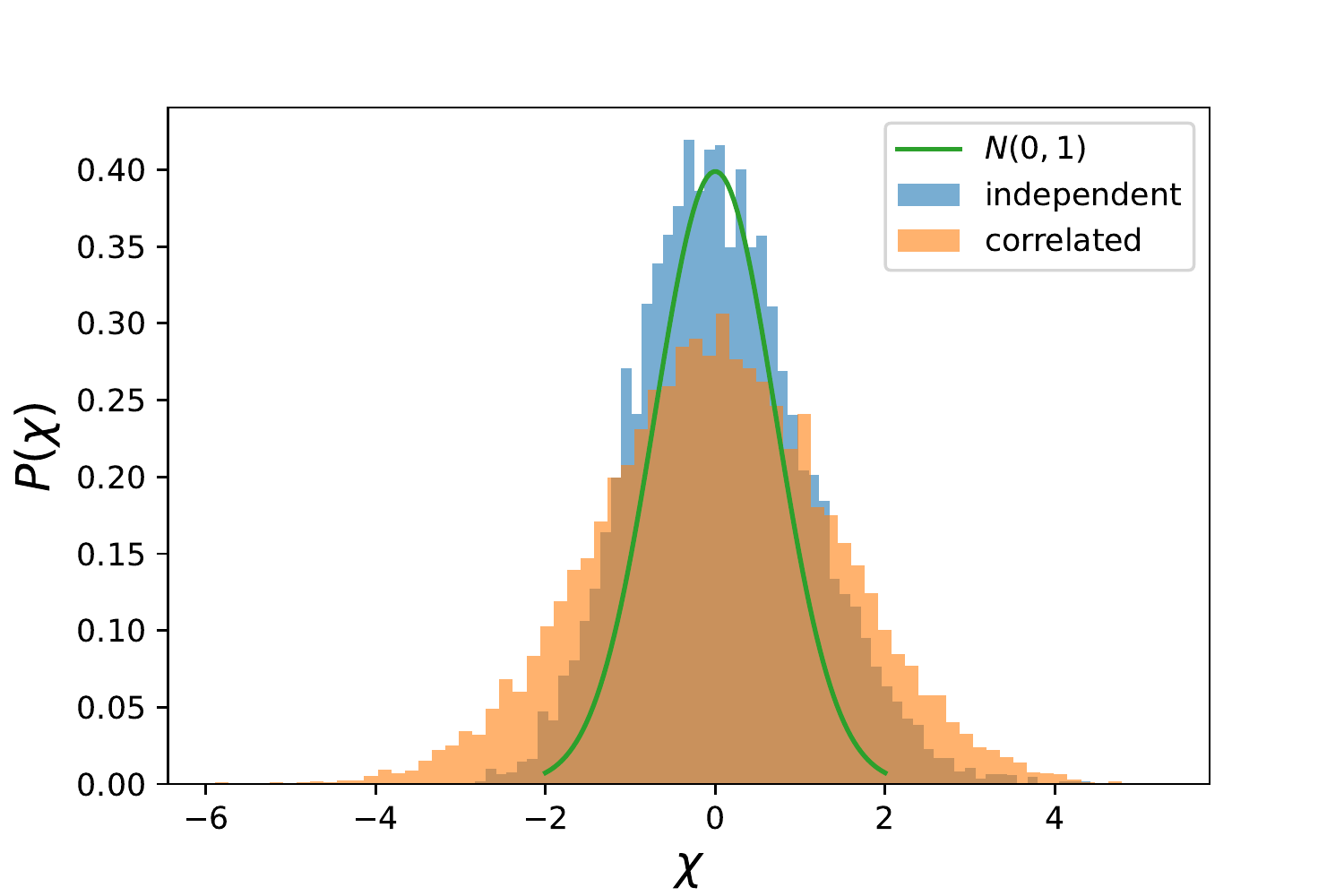}
    \caption{The pool distributions (see text) show that the correlation between the neighbourhoods of the points brings to an error estimate which slightly underestimates the value that would be obtained using statistically independent samples.}
    \label{fig:pool}
\end{figure}

To begin with, we generated 10000 realizations of 10000 uniformly distributed points on a 4-dimensional lattice. For each realization, we extracted ID and error using all available points; on the same dataset, we also computed the ID using only one random point, in order to gather statistically independent measurements. We then computed the distribution of the pool variable for the correlated ID measures as
\begin{equation}
    \chi_{corr} = \frac{d_{cor}-d_{gt}}{\sigma_{Bayes}},
    \label{eq:pool_all}
\end{equation}
where $\sigma_{Bayes}$ is the standard deviation of the posterior, and $d_{gt}$ is the ground truth ID. We also computed the distribution of the pool variable for the ID estimates obtained using a single point 
\begin{equation}
    \chi_{ind} = \frac{d_{ind}-d}{\sigma_{stat}},
    \label{eq:pool_ind}
\end{equation}
where $\sigma_{stat}$ is the standard deviation of the distribution of the single-point ID estimates. We expect $\chi_{ind}\sim\mathcal{N}(0,1)$ and that's what we obtained as shown from the blue histogram in Fig.~\ref{fig:pool}. On the other hand, the distribution obtained for the pool of correlated measurements $\chi_{cor}$ (orange histogram in Fig.~\ref{fig:pool}) shows a higher spread, indicating that the $\sigma_{Bayes}$ systematically underestimate the error of the $\sim30\%$. 
This was expected, as both the Bayesian and likelihood formulation assume to sample statistically independent observations.
We can then conclude that the statistical dependence of neighborhoods in the calculation of ID leads to an error estimate which is slightly below the correct value but it is still very indicative.

\section{Analytical results in continuum space}
\label{sec:continuum}
The ID estimation scheme proposed in our work can be easily extended and applied to different metrics than the lattice one  (used to build the I3D).
In particular, within any $L^p$ metric in the continuum space, the volume of the hyper-spheres scales as a canonical power law of the radius multiplied by a prefactor depending on both the dimension $d$ and the value $p$:
\begin{equation*}
V_d^p(R)= \frac{(2\Gamma(\frac{1}{p}+1)R)^d}{\Gamma(\frac{d}{p}+1)} = \Omega_d^pR^d.    
\end{equation*}
As a consequence, the ratio of volumes that occurs in the binomial distribution of eq.~\eqref{eq:region_ratio} becomes
\begin{equation}
    p(r,d)=\frac{\Omega_dR_{1}^d}{\Omega_dR_{2}^d}=\left(\frac{R_{1}}{R_{2}}\right)^d\!\!=r^d .
    \label{eq:volumes_ratio_short}
\end{equation}
Because of the well-behaved scaling of the volume with the radius in continuous metrics, all formulas presented so far for the discrete case consistently simplify, allowing for further analytical derivations concerning both the MLE and Bayes analyses, as shown in the two following sections.
\subsection{Maximum Likelihood Estimation and Cramer-Rao lower bound}
In the continuum case, given the previous expression for the ratio of volumes, the MLE and Cramer-Rao relations can be utterly simplified from Eq.~\eqref{eq:binom_mle} so that we can obtain an explicit form for the intrinsic dimension. Concretely, by substituting $p=r^d$ into eq.~\eqref{eq:binom_mle} we find
\begin{equation}
    d=\frac{\ln(\E n / \E k)}{\ln(r)}
    \label{eq:d_est}
\end{equation}
for the MLE, while the Cramer Rao inequality~\eqref{eq:CR_var} becomes
\begin{equation}
    \text{Var}(d;r,N,k) \ge \frac{1-r^d}{\E k N\ln(r)^2 r^d} \;.
    \label{eq:CR_cont}
\end{equation}
In order to have an estimate as precise as possible, we are interested in the value of $r$ that minimize the variance. Being~\eqref{eq:CR_cont} a convex function, we find a single minimum that corresponds to
\begin{equation}
    r_{opt}(d) = 2^{-1/d} \left(-W\left(-\frac{2}{e}\right)\right)^{1/d}\! \approx\; 0.2032^{\frac{1}{d}}
    \label{eq:r_opt}
\end{equation}
where $W$ is the Lambert $W$ function. Of course in principle we don't know the intrinsic dimension of the system, and thus we don't have a direct way to practically select $r_{opt}$ if not through successive iterations. This relationship tells us that higher $d$ needs higher $r$ to provide a better estimate. 
The above relation also suggests that the optimal ratio between the points in the two shell should approach $\E n / \E k\sim r_{opt}^d=0.2032$. This implies that there is an optimal and precise fraction of points, and thus volumes, for which the estimated ID is more accurate.
Supposing that we are able to find such $r_{opt}$, we might ask how the variance scales with the dimensionality of the system. We obtain that
\begin{equation}
    \text{Var}(d;r_{opt},N,k)=\frac{1-r_{opt}^d}{\ln(r)^2 kNr_{opt}^d} \propto \frac{d^2}{Nk} 
    \label{eq:var_opt}
\end{equation}
This implies that the precision of our estimator scales with the square of the dimension.
\subsection{Bayes formulation}
Also the Bayesian derivations bring to analytical results in the continuum case. In particular, from eq.~\eqref{eq:posterior}, one obtains
\begin{equation}
    p(d) =\text{Beta}(r^d;\alpha,\beta)|r^d\ln(r)|.
    \label{eq:post_d}
\end{equation}
From this expression one can easily derive the first and second momenta of the distribution. In particular, performing the change of variable $d = \ln{x}/\ln{r}$, we have
\begin{equation}
    \E{d} = \int_0^\infty \text{d}d \; p(d)d = -\frac{1}{\ln(r)}\int_0^1 \text{d}x\;\text{Beta}(x;\alpha,\beta)\ln(x)=\frac{\psi_0(\alpha)-\psi_0(\alpha+\beta)}{\ln(r)}=
    \frac{\psi_0(1+\sum_{i=1}^Nn_i)-\psi_0(2+\sum_{i=1}^Nk_i)}{\ln(r)}
\end{equation}
where $\psi_0(z)=\frac{\text{d}}{\text{d}z}\ln{\Gamma(z)}$ is the digamma function; in the last step we have inserted the definitions for $\alpha$ and $\beta$ from Eq.~\eqref{eq:beta_par}.
By exploiting the same change of variable, also the variance ends up into a simple expression:
\begin{equation}
    \text{Var}(d) = \frac{\text{Var}(\ln(x))}{\ln(r)^2} = \frac{\psi_1(\alpha)-\psi_1(\alpha+\beta)}{\ln(r)^2} =
    \frac{\psi_1(1+\sum_{i=1}^Nn_i)-\psi_1(2+\sum_{i=1}^Nk_i)}{\ln(r)^2}
\end{equation}
where $\psi_1(z)=\frac{\text{d}^2}{\text{d}z^2}\ln{\Gamma(z)}$ is the trigamma function.

\section{Fractal lattices}
\begin{figure}[htb]
    \centering
    \includegraphics[width=0.9\linewidth]{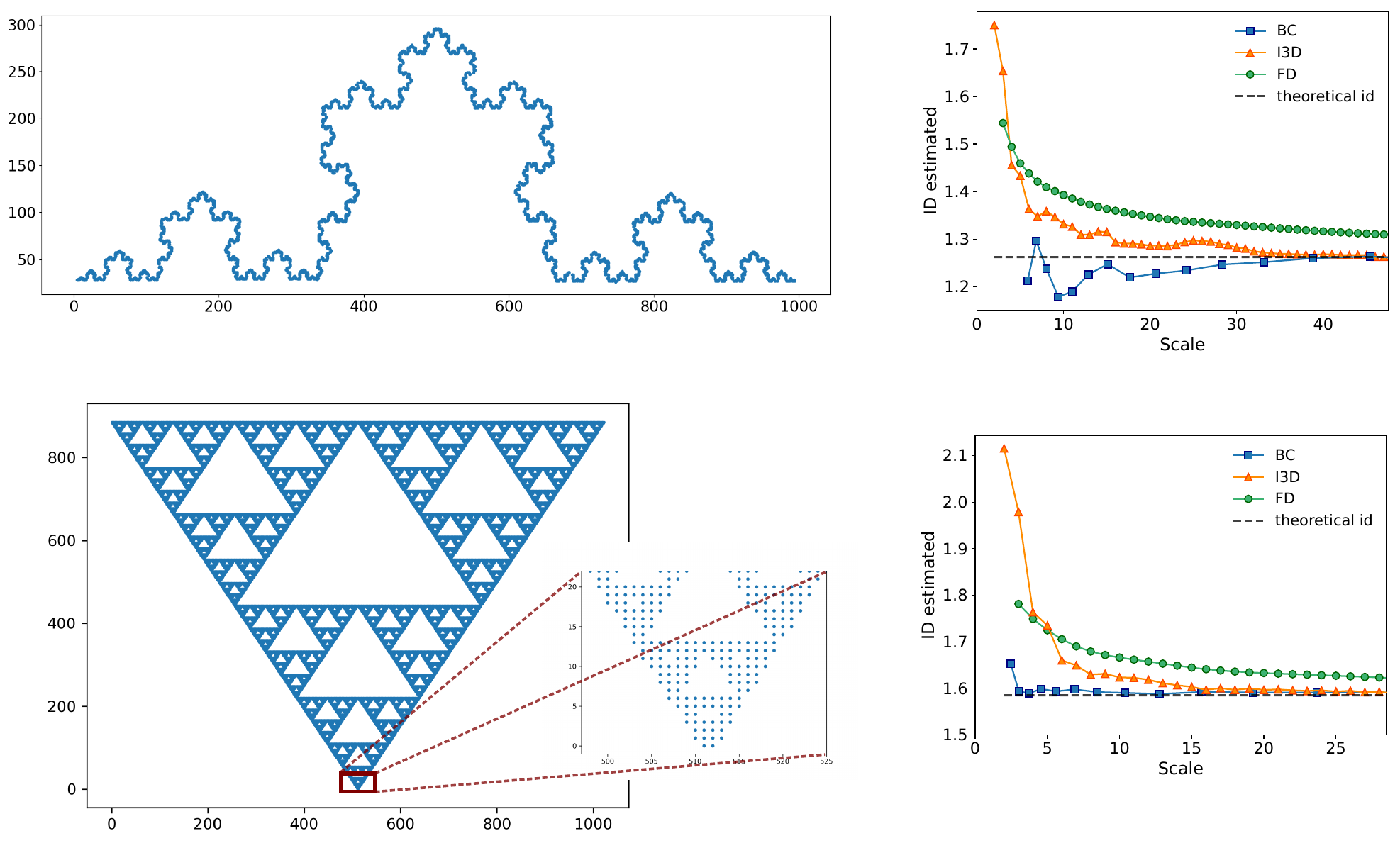}
    \caption{I3D, BC and FD are all capable of finding the correct theoretical ID for fractal lattices, even if that occurs on different scales.}
    \label{fig:SI_frac}
\end{figure}

As a further test to check the goodness of I3D, we compared the performance of the estimators on discrete fractal lattices, where Box Counting (BC) and Fractal Dimension (FD) already proved to be reliable\cite{falconer2004fractal,niemeyer1984fractal}. In Fig.~\ref{fig:SI_frac}, we report the ID as a function of the scale for the Koch curve (above), whose ID is $\log(4)/\log(3)\sim1.26$ and the Sierpinski gasket (below), with an ID of $\log(3)/\log(2)\sim1.58$.
As one can appreciate, the FD converges to the proper values only for large scales, where the discrete nature of the dataset is negligible, while BC and I3D quickly find the correct ID.
However, the I3D adds a piece of information: it clearly shows at which scale the "fractality" of data comes into play. Indeed, by looking at the inset of the Sierpinski gasket, one notices that at a scale smaller than five the 2-dimensional structure is still prevailing.

\section{Spin systems}
\begin{figure}[htb]
    \centering
    \includegraphics[width=0.75\linewidth]{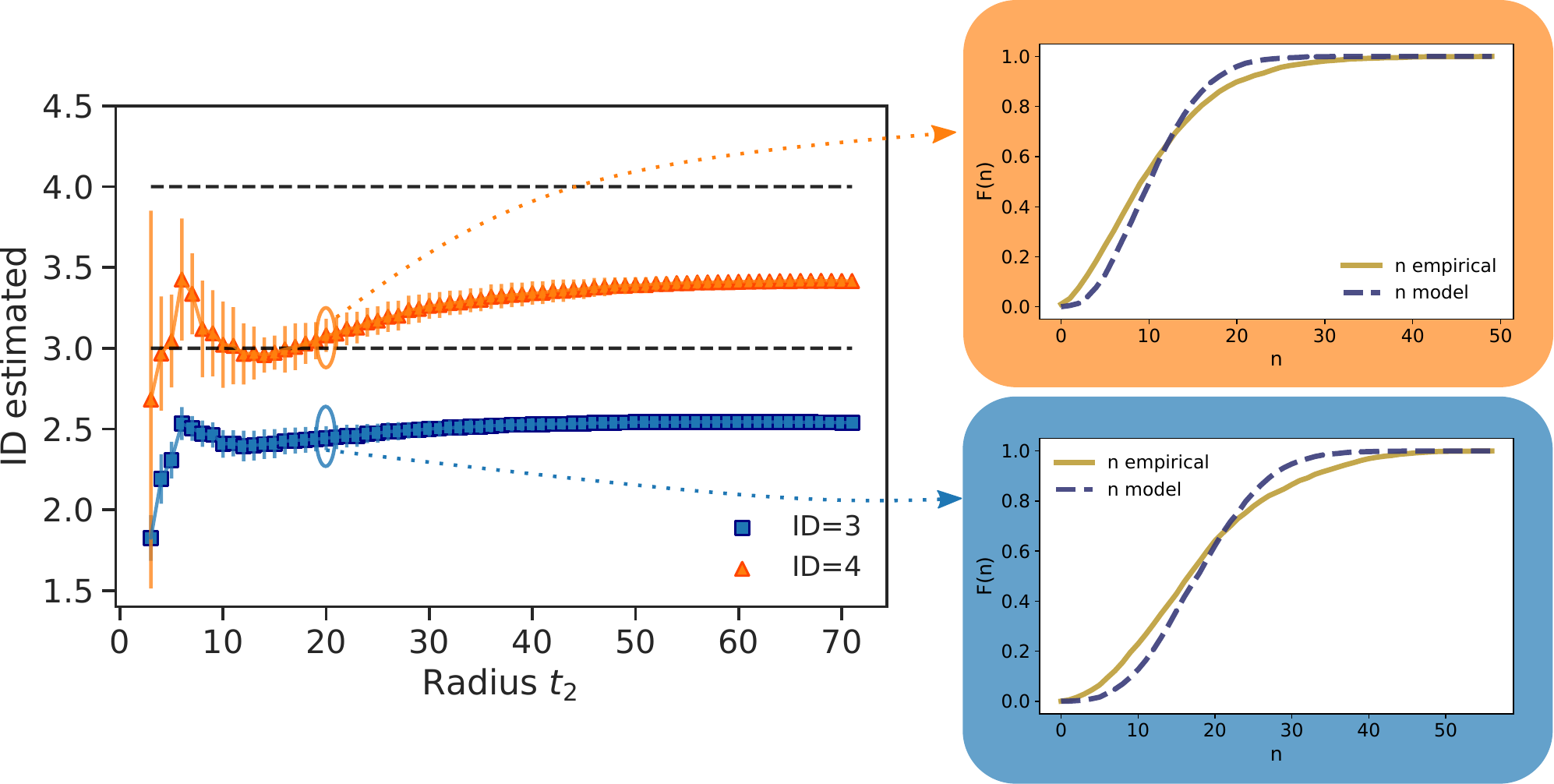}
    \caption{The I3D estimator underestimates the ground truth ID of spin systems in the case of $d=3$ and $d=4$. The inconsistency is shown by the model validation plots, where theoretical and empirical cdf are not well superimposed.}
    \label{fig:SI_spins}
\end{figure}
\subsection{PCA on discrete spins}
Here we address the possibility of performing PCA on discrete datapoints, justifying the results of the last section of the main text.
We start from recalling that the continuous spins were generated using a linear embedding $\bm{\varphi}_i=\bm{\varphi}_0+\bm\alpha\epsilon(i)$, so that it is possible to retrieve the directions of the generating vectors $\bm\alpha_i$ using standard PCA and a number of points $N\ge \text{ID}+1$.

In the case of spin states, the retrieval of $\bm \alpha$ is not so straightforward.
In particular, two spin states differ from each other by a finite (and possibly very small) amount of spin flips. This means that we have a piece of information only on a fraction of the features, namely the varying spins. 
For this reason, we need many more points in order to gather statistics about the behaviour of the spins and how often they flip across the dataset. The idea is that PCA eigenvectors can capture the frequency of spin flips and give a proxy of the embedding directions $\bm\alpha_i$.
The result for ID=1 is reported in Fig.~9 and compares the generating vector $\bm\alpha$ and the first PCA eigenvector $\bm{v}_1$. The overlap is almost perfect, as we have $\bm\alpha\cdot\bm{v}_1\sim0.98$.
\begin{figure}[htb]
    \centering
    \includegraphics[width=0.85\linewidth]{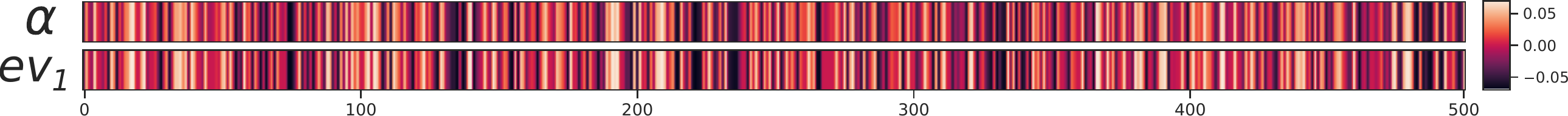}
    \caption{Comparison of the embedding vector $\bm\alpha$ and the first PCA eigenvector $\bm{v}_1$ obtained with $\sim1000$ points in the 1-dimensional case. The overlap is almost perfect: $\bm\alpha\cdot\bm{v}_1\sim0.98$.}
    \label{fig:SI_spin_PCA}
\end{figure}
In higher dimensions, the eigenvectors will be rotated with respect to the original embedding vectors, so that a direct visual comparison cannot be made as in the previous case. Hence,  we estimate  the residual of the overlap, defined as
\begin{equation}
    \mathcal{R}=d-\sum_{i,j=1}^d (\bm\alpha_i\cdot\bm{v}_i)^2.
\end{equation}
In the 2-dimensional case, we find that $\mathcal{R}\sim0.04$, meaning that, like for the 1-dimensional example, we are able to retrieve $\sim98\%$ of the generative process information. 

\section{Results for different nucleotide sequence distances and choices of radii ratio}
\begin{figure}[b]
    \centering
    \includegraphics[width=0.75\linewidth]{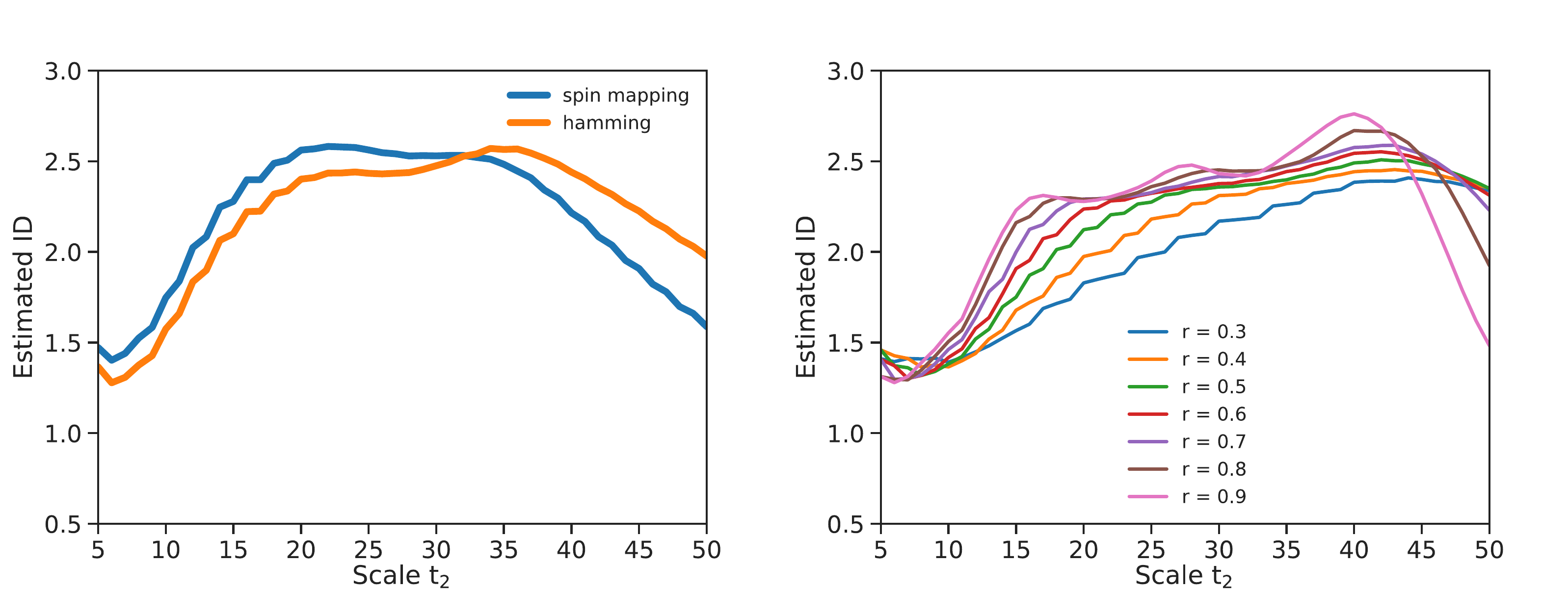}
    \caption{Different choice of mapping (left) or parameter $r$ (right) only slightly affect the ID estimate.}
    \label{fig:map_one}
\end{figure}
One might wonder how the ID estimate depends on the metric used to compute the distance between sequences. As, anticipated in the main text, we considered two possible mappings. 
The first one maps each letter to a two spin state as follows:
A:11, T:00, C:10 and G:01. The distance is then measured through Hamming or Manhattan indifferently. As a consequence complementary purine and pyrimidine are at distance 2, while other distances are just 1. The other possibility is to use the Hamming distance straightly on the sequences as they are, meaning that all nucleotides are equidistant one from the other.
Fig.~10 (left) shows the ID as a function of the scale of the same cluster used in the main text. The behaviour of the ID depends very slightly on the chosen distance measure. For this reason we decided to stick to the spin mapping, as it allows retrieving the local "directions" of the generating process by a PCA analysis.
Seemingly, different choices of the free parameter $r=t_1/t_2$ do not noticeably affect the ID estimate, especially in the plateau region ($15<t_2<40$), where an ID can thus properly defined.  